\documentclass[runningheads]{llncs}
\usepackage[T1]{fontenc}
\usepackage{cite}
\usepackage{amsmath,amssymb,amsfonts}
\usepackage{algorithmic}
\usepackage{graphicx}
\usepackage{textcomp}
\usepackage{todonotes}
\usepackage{multirow}
\usepackage{xcolor}
\usepackage{orcidlink}
\usepackage{url}
\usepackage{booktabs}
\usepackage{bm}
\usepackage{float}

\begin{document}
\title{Leveraging Data Augmentation and Siamese Learning for Predictive Process Monitoring}
\titlerunning{Leveraging Data Augmentation and Siamese Learning for PPM}
%
\author{Sjoerd van Straten\inst{1}\orcidlink{0009-0002-5772-9073} \and
Alessandro Padella\inst{2}\orcidlink{0009-0008-6399-9364} \and
Marwan Hassani\inst{1}\orcidlink{0000-0002-4027-4351}}
\authorrunning{van Straten et al.}
%

\institute{
  Department of Mathematics and Computer Science,
  Eindhoven University of Technology, The Netherlands\\
  \email{\{h.a.j.v.straten, m.hassani\}@tue.nl}
  \and
  Department of Mathematics and Computer Science,
  University of Padua, Italy\\
  \email{alessandro.padella@unipd.it}
}

\maketitle              
\begin{abstract}
Predictive Process Monitoring (PPM) enables forecasting future events or outcomes of ongoing business process instances based on event logs. However, deep learning PPM approaches are often limited by the low variability and small size of real-world event logs. To address this, we introduce SiamSA-PPM, a novel self-supervised learning framework that combines Siamese learning with Statistical Augmentation for Predictive Process Monitoring. It employs three novel statistically grounded transformation methods that leverage control-flow semantics and frequent behavioral patterns to generate realistic, semantically valid new trace variants. These augmented views are used within a Siamese learning setup to learn generalizable representations of process prefixes without the need for labeled supervision. Extensive experiments on real-life event logs demonstrate that SiamSA-PPM achieves competitive or superior performance compared to the SOTA in both next activity and final outcome prediction tasks. Our results further show that statistical augmentation significantly outperforms random transformations and improves variability in the data, highlighting SiamSA-PPM as a promising direction for training data enrichment in process prediction.

\end{abstract}
\section{Introduction}\label{sec:introduction}
\begin{figure}[ht]
    \centering
    \includegraphics[width=0.7\linewidth]{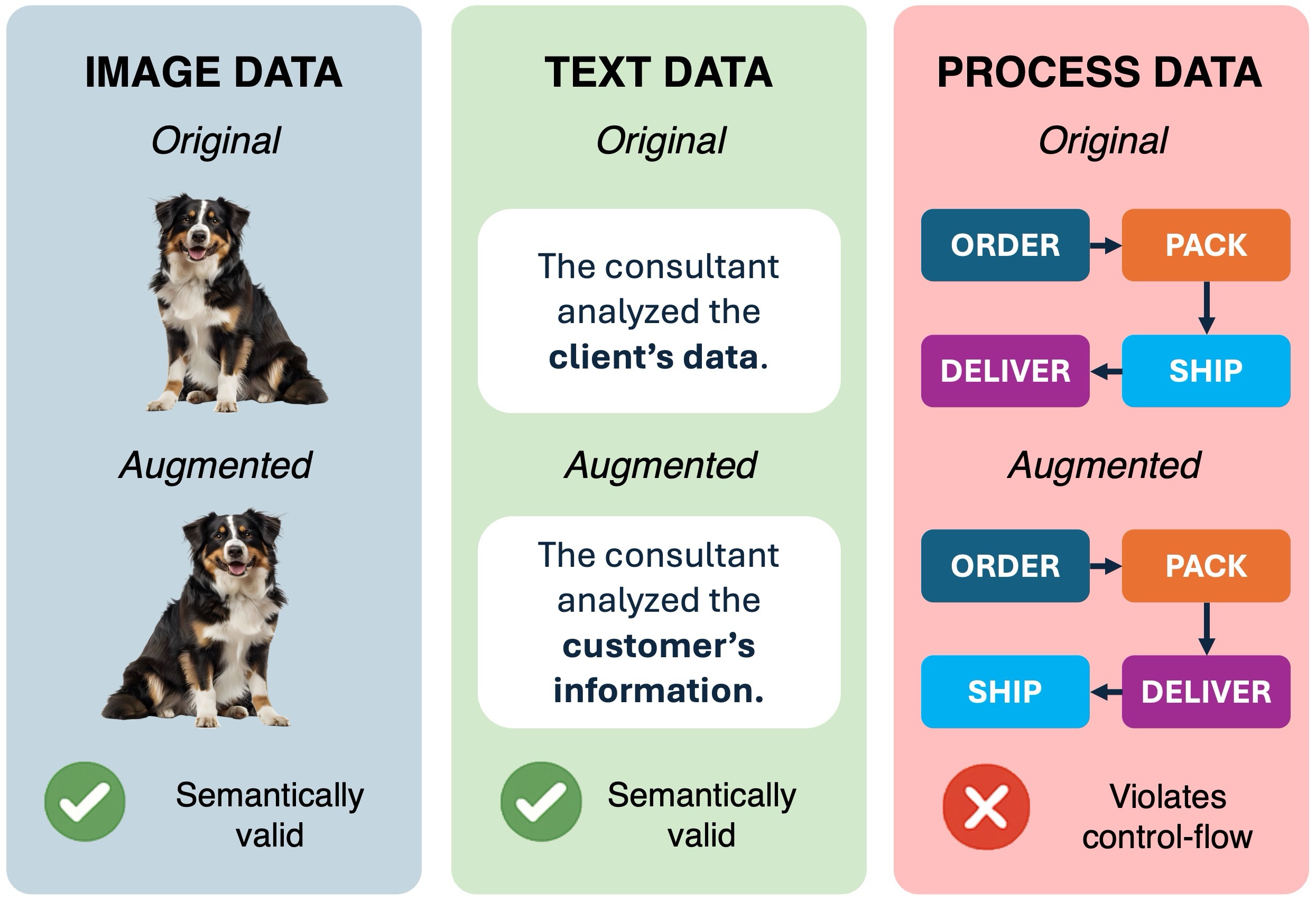}
    \caption{ Random augmentation strategies can preserve semantics in image and text domains, where structural flexibility exists. However, in process data, such augmentations risk violating control-flow dependencies for instance, producing logically inconsistent or non-executable sequences.}
    \label{fig:introduction_image}
\end{figure}

Due to the increasing digitization of business operations, Predictive Process Monitoring (PPM) has become a critical area of research within process mining. Two of its core tasks, \textit{next activity prediction} and \textit{final outcome prediction}, have received substantial attention due to their practical relevance in optimizing workflows, reducing delays, and improving resource utilization \cite{Verbeek2025}. While earlier works utilized symbolic models such as Petri nets and transition systems, recent advancements in deep learning, especially sequence models such as Long-Short Term Memory (LSTM), Gated Recurrent Unit and attention-based Transformers, have significantly improved the predictive performance of such models. However, a persistent challenge in deep learning and inherently in its application to PPM, is the limited size of available labeled data \cite{ceravolo2024predictive}. Despite their high complexity, business processes are typically logged with limited trace variation and often result in datasets containing redundant or highly similar samples. These limitations affect the generalizability of predictive models, especially in scenarios with infrequent behaviors or class imbalance. In machine learning, a common remedy for such constraints is \textit{data augmentation}: the process of artificially generating additional training data to enhance diversity and robustness. Data augmentation has been extensively explored in domains such as computer vision and natural language processing (NLP) \cite{wang2024comprehensive}. In these fields, augmentations are informed by domain knowledge to preserve the semantic integrity of the data. For example, visual enhancements such as flipping or cropping retain object identity, while textual methods such as synonym replacement, word deletion or insertion aim to maintain grammatical and contextual coherence~\cite{wei2019eda}. The core principle in these domains is that augmentations should be \textit{semantically valid} and \textit{task-relevant}. In contrast, data augmentation for process mining remains considerably underexplored. Prior work explored model-agnostic augmentation using primarily random insertions, deletions or replacements \cite{kappel2023augmentation}, but such random transformations risk violating process semantics. For example, delivering an order before it has been shipped (cf. Fig.~\ref{fig:introduction_image}). In contrast, our approach introduces \textbf{statistically grounded augmentation techniques} that leverage frequent control-flow patterns to generate realistic, semantically valid trace variants, thereby aligning augmentation with actual process dynamics (cf. Section~\ref{sec:related_work}). We use the process in Fig.~\ref{fig:introduction_image} $(Order \rightarrow Pack \rightarrow Ship \rightarrow Deliver)$ as a running example throughout the paper to illustrate how random augmentations can violate semantics, while our statistically grounded methods preserve realistic control-flow.

In addition to augmentation, we also explore the underutilized potential of \textit{self-supervised learning} (SSL) in PPM. While most existing approaches rely on supervised learning, this requires labeled data (e.g. next activity or final outcome annotations) for each trace. Practically, such amount of labels may not be sufficient to train a model. In fact, some event logs may be imbalanced, resulting in, for example, an insufficient number of examples for a given outcome, e.g. a given machine breaks down. This issue also applies to the next activity prediction task, where the occurrence of certain activities may be rare, leading to a very limited number of events with the corresponding label. SSL addresses this by learning from the data itself without requiring manual labels. It constructs surrogate tasks, such as aligning different augmented views of the same input to learn robust task-agnostic representations. Although SSL has seen great success in the computer vision domain \cite{chen2020simclr, grill2020byol, chen2021simsiam}, it is underexplored in PPM. In this work, we adopt a \textbf{Siamese learning} strategy based on Bootstrap Your Own Latent (BYOL) \cite{grill2020byol}, where augmented trace prefixes are encoded to similar latent representations. This setup avoids the need for negative samples or large batch sizes \cite{grill2020byol}, making it well-suited for structured, small-scale process logs. Combined with our statistically grounded augmentations, this enables high-quality representation for downstream tasks, even with limited labeled data.

The contributions of this paper include: i) introducing three novel, statistically grounded trace transformation methods that preserve behavioral semantics and process structure, ii) introducing a self-supervised Siamese framework that learns robust embeddings of process prefixes using augmented trace pairs, iii) conducting extensive experiments on eight real-world case studies, demonstrating the performance of our approach in both next activity prediction and final outcome prediction and iv) comparing our framework against four state-of-the-art baselines, analyzing predictive performance and computational efficiency.

The remainder of this paper is organized as follows. Section \ref{sec:related_work} discusses related work in data augmentation, self-supervised learning and PPM. Section \ref{sec:problem_definition} formally defines the problem setting. Section \ref{sec:method} presents our proposed method, \textit{SiamSA-PPM}, including the augmentation strategy and the SSL framework. Next, Section \ref{sec:experiments} reports on our experimental setup, after which Section~\ref{sec:results} shows our results with an associated ablation study. Finally, Section \ref{sec:conclusion_and_future_work} concludes with directions for future work.


\section{Related Work}\label{sec:related_work}

\subsubsection{Data Augmentation Techniques}\label{subsec:data_augmentation}
Data augmentation is a proven strategy to increase training data and improve model generalization. In NLP, a well-known method is EDA \cite{wei2019eda}, which introduces random augmentations such as random insertion, deletion and replacement of words. Other techniques such as syntax-aware transformations \cite{duan2023syntax} and generative approaches such as InsNet \cite{lu2022insnet} inject linguistic diversity into text classification and translation tasks. In computer vision, simple pixel-level augmentations and mixing techniques (e.g. ReMix \cite{chou2020remix} and SmoothMix \cite{lee2020smoothmix}) enhance model robustness, while more recent work introduces frequency- and structure-based modifications \cite{hwang2023improving}. These augmentation strategies have proven highly effective in both supervised and self-supervised learning. In contrast to NLP and computer vision, data augmentation in PPM remains underexplored. The work in~\cite{kappel2023augmentation} introduced the first comprehensive model-agnostic augmentation framework for event logs. Their method applies a set of timestamp-preserving transformations, including random insertion, deletion, replacement and swap (derived from \cite{wei2019eda}), but also introduces swapping of events with the same timestamp and loop augmentation. Although these transformations increase variability, a key limitation is that they are partly applied at random, without considering the underlying process structure. This can result in unrealistic behavior that contradicts the actual control-flow semantics of the process. As such, while the method enriches the training data, it does not guarantee that the new traces are useful or even likely within the process domain. This gap motivates the development of \textbf{statistically grounded transformations}, which are transformations based on frequent paths or XOR-splits mined from the original training data. Such augmentations can maintain control-flow while still introducing meaningful variability to support generalization. 

\subsubsection{Self-Supervised Learning (SSL) Approaches}\label{subsec:ssl_approaches}
\begin{figure}[h]
    \centering
    \includegraphics[width=0.8\linewidth]{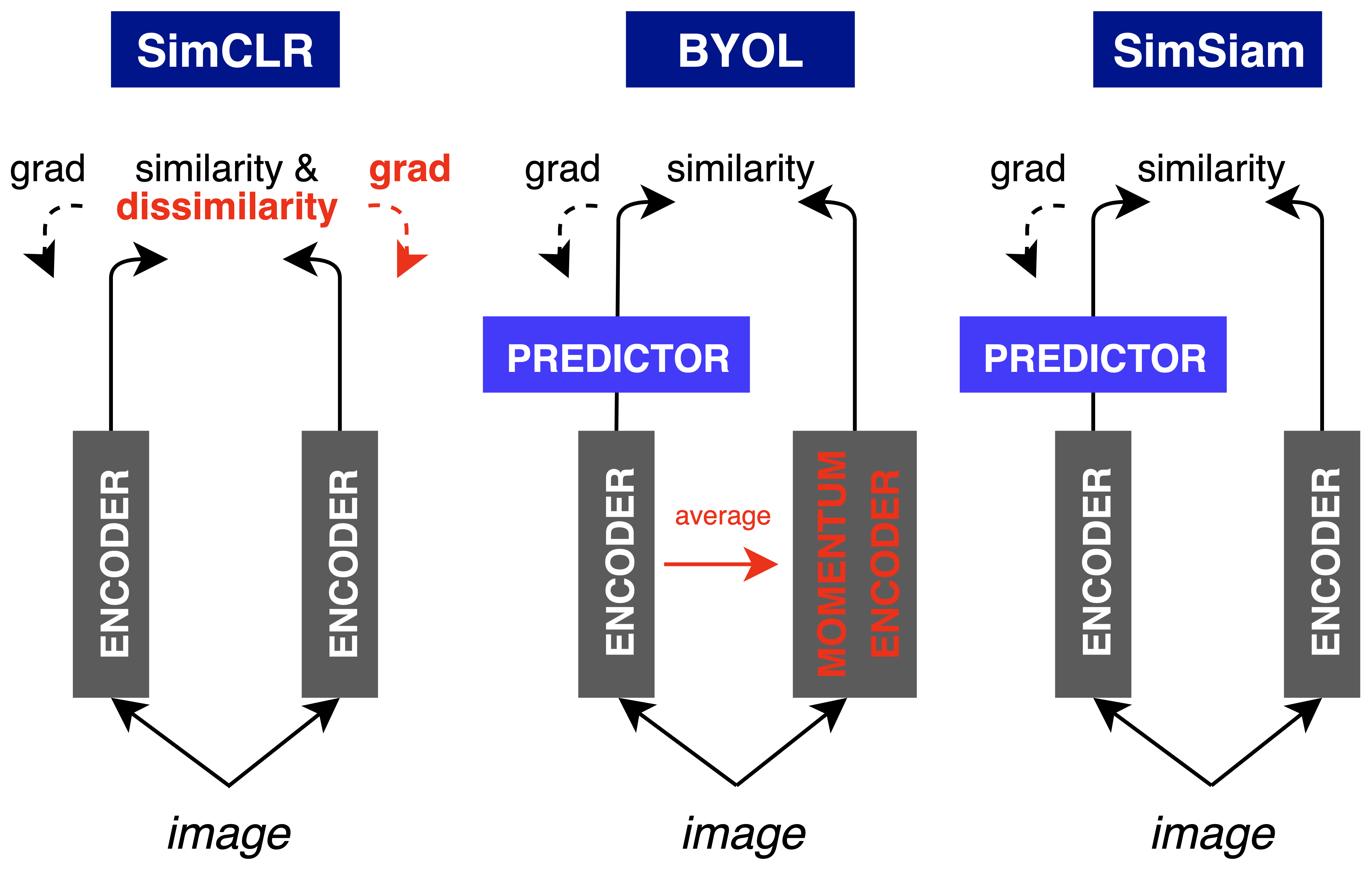}
    \caption{Overview of three self-supervised learning frameworks. SimCLR \cite{chen2020simclr} uses contrastive learning with negative pairs, BYOL \cite{grill2020byol} introduces a momentum encoder and avoids negatives, while SimSiam \cite{chen2021simsiam} simplifies the setup using only a stop-gradient operation.}
    \label{fig:ssl_methods}
\end{figure}

have emerged as powerful tools to learn useful representations from unlabeled data. Instead of relying on manual annotations, SSL defines pretext tasks that use the data itself to generate training input. This has proven particularly effective in domains where labeled data is scarce or expensive to obtain. A dominant class of SSL methods uses contrastive learning techniques. SimCLR \cite{chen2020simclr} generates two augmented views of the same input and trains the model to bring these closer in the embedding space while pushing apart other (negative) samples. While effective, contrastive methods require careful design of negative pairs, large batch sizes and heavy data augmentation. This makes them difficult to apply to structured, small-scale domains such as PPM. To overcome these limitations, \textbf{Siamese learning} frameworks like BYOL \cite{grill2020byol} and SimSiam \cite{chen2021simsiam} have gained popularity (cf.\ Fig.~\ref{fig:ssl_methods}). These models use two identical networks to process different augmented versions of the same input, learning to align their latent representations. Unlike contrastive methods, they do not require negative pairs or large batch sizes. BYOL introduces a momentum-updated target network, while SimSiam claims to achieve similar performance with simpler architectures and a stop-gradient mechanism \cite{chen2021simsiam}. Both frameworks have shown strong results in learning high-quality embeddings across NLP and computer vision tasks. 

SSL enables representation learning from unlabeled data, which can benefit PPM tasks such as next activity prediction and final outcome prediction. Pre-training with SSL can also help when labeled traces are limited or imbalanced, which is often the case, especially for final outcome prediction. Moreover, SSL encourages the model to learn structural and semantic properties of the data, rather than overfitting to specific task labels. However, the quality of learned representations is clearly dependent on the diversity and the relevance of the augmentations used to create positive pairs. This makes statistically grounded data augmentation not just beneficial, but essential for effective representation learning in PPM.

\subsubsection{Offline Predictive Process Monitoring (PPM) Approaches}\label{subsec:ppm_approaches}
aim to forecast the future behavior of ongoing process instances such as the next activity, remaining time or final outcome, based on event log data. While early approaches relied on formal models (e.g. Petri nets and transition systems), deep learning models now dominate due to their ability to learn complex patterns~\cite{ceravolo2024predictive}. In particular, models such as LSTMs \cite{tax2017predictive} and Convolutional Neural Networks (CNNs) \cite{dimauro2019activity, pasquadibisceglie2019cnn} have been widely used for PPM. More recent models have leveraged attention mechanisms and Transformer architectures for improved performance \cite{bukhsh2021processtransformer}, while others explored autoencoders \cite{ni2022predicting} and adversarial frameworks \cite{taymouri2020predictive,10.1007/978-3-031-81375-7_21}. Despite these advances, a general issue with deep learning models is that they require large datasets to generalize well (cf. Section~\ref{sec:introduction}). To ensure stable training, it is recommended to have at least one unique training sample for each parameter being estimated \cite{goodfellow2016deep}. As deep learning models estimate thousands or even millions of parameters during training and event logs generally contain only a few thousand traces, often with redundant samples, there is a need for an artificial increase of training data.

\section{Problem Definition}\label{sec:problem_definition}

Let $\mathcal{L} = \{ \sigma_1, \sigma_2, \dots, \sigma_N \}$ denote an event log composed of $N$ traces, where each trace $\sigma_i = \langle e_{i,1}, e_{i,2}, \dots, e_{i,T_i} \rangle$ is a sequence of $T_i$ events associated with a process instance. Each event $e_{i,t}$ carries attributes such as the activity label, timestamp and possibly additional case or event-level features. Each event $e_{i,t}$ is here intended as a tuple $(A_{i,t}, t_{i,t}, \vec{f}_{i,t})$, in which the activity $A_{i,t}\in\mathcal{A}$ for $\mathcal{A}$ the set of possible activities of the event log $\mathcal{L}$,  $t_{i,t}\in\mathcal{T}$ the set of admitted timestamps and $\vec{f}\in\mathcal{F}_1\times\ldots\times\mathcal{F}_m$, where $\mathcal{F}_1,\ldots,\mathcal{F}_m$ are the sets containing the values of the different features. To lighten the notation, in this paper we are going to refer to $\langle A_1, \ldots, A_n \rangle$ as the sequence of events $\langle e_1, \ldots, e_n \rangle $ for which the corresponding activities are $A_1,\ldots, A_n$ and both the timestamps and the feature vectors are not known. In line with common practice in PPM, we treat each recorded event as the completion of an activity. Now, it is possible to define the prediction problems as follows:

\begin{enumerate}
    \item \textbf{Next Activity Prediction:} Given $\sigma^{(k)}_i$, predict $\hat{A}_{k+1,t}$ that should ideally match the ground truth activity label $A_{k+1,t}$.

    \item \textbf{Final Outcome Prediction:} Given $\sigma^{(k)}_i$, predict if there exists an event $e_{\hat{k}, i}\in\langle e_{k+1,i},\ldots,e_{n,i} \rangle$ such that $e_{\hat{k}, i}=(t, \hat{A}_{out}, \vec{f})$ for some $t$ and $\vec{f}$ where $\hat{A}_{out}$ should ideally match $A_{out}$ the ground truth final outcome label.
\end{enumerate}   

A major challenge in this setting is the scarcity of labeled event logs, which limits the ability to train accuracy predictive models \cite{ceravolo2024predictive}. Therefore, it is necessary to enrich the event logs with \textit{meaningful} instances to improve predictive performance.
\section{Our Method}\label{sec:method}

In this section, we present our methodology for event log augmentation and pretraining a robust encoder for the PPM tasks introduced in Section~\ref{sec:problem_definition}. In Section~\ref{subsec:novel_transformation_methods} we first introduce three statistically grounded trace transformation techniques based on activity sequences, while Sections~\ref{subsec:data_augmentation_strategy} and~\ref{subsec:siamese_pre_training} outline the different data augmentation pipelines and Siamese pretraining strategy, respectively. Finally, the fine-tuning procedure is introduced in Section~\ref{subsec:fine_tuning}.

\subsection{Novel Transformation Methods}\label{subsec:novel_transformation_methods}
\begin{figure}[h]
    \centering
    \includegraphics[width=0.8\linewidth]{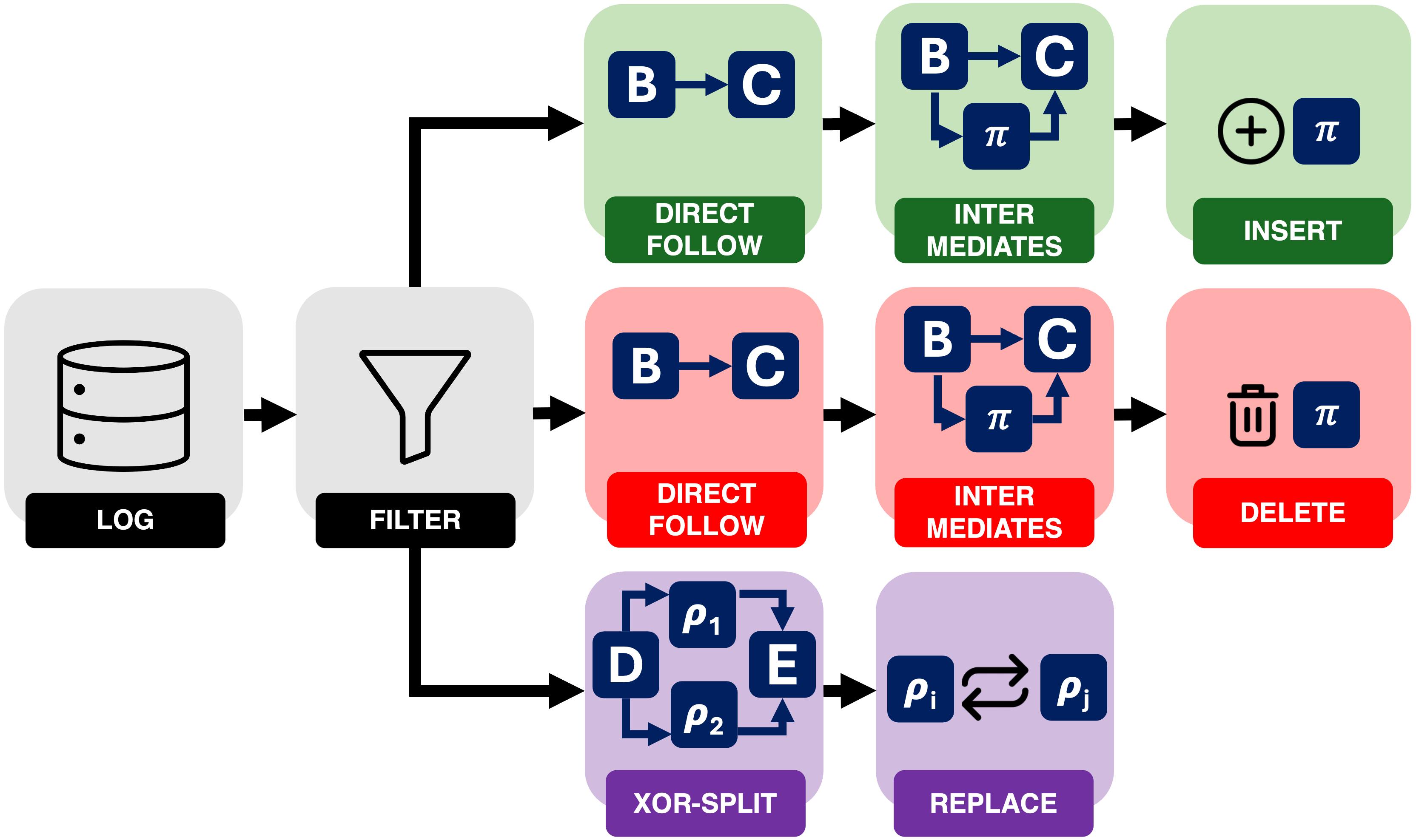}
    \caption{Overview of our three novel transformation methods: StatisticalInsertion (green), StatisticalDeletion (red) and StatisticalReplacement (purple).}
    \label{fig:novel_transformations}
\end{figure}

To keep the semantics of process traces and effectiveness of data augmentation in PPM, we propose three statistically grounded transformation techniques, named \textbf{StatisticalInsertion}, \textbf{StatisticalDeletion} and \textbf{StatisticalReplacement} (cf. Fig.~\ref{fig:novel_transformations}). Unlike traditional random transformations, these methods leverage frequent activity patterns in the event log to generate realistic and process-compliant trace variants. The techniques are parameterized by thresholds $\alpha, \beta, \gamma, \delta \in [0,1]$ and a maximum intermediate length $\lambda_{\max}$, which control frequency constraints and structural complexity.

Both \textit{StatisticalInsertion} and \textit{StatisticalDeletion} operate on observed direct and intermediate activity sequences. The procedure comprises the following steps. \textbf{1) Activity filtering.} From the original training log $\mathcal{L}$, we construct a filtered log $\mathcal{L}' \subseteq \mathcal{L}$ by retaining only traces composed of activities that occur in at least factor $\alpha$ of all cases. This is solely needed for preprocessing and depends on the associated prediction task. \textbf{2) Direct follower extraction.} From $\mathcal{L}'$, we extract all frequent direct follower pairs $(B \rightarrow C)$ where activity $C$ directly follows $B$ in at least factor $\beta$ of all observed transitions. These pairs represent commonly co-occurring dependencies. \textbf{3) Intermediate sequence mining.} For each direct pair $(B \rightarrow C)$, we identify frequent intermediate subsequences $\pi = \langle A_1, \dots, A_k \rangle$ such that $B \rightarrow \pi \rightarrow C$ occurs in at least factor $\gamma$ of traces in $\mathcal{L}'$, with $1 \leq k \leq \lambda_{\max}$. \textbf{4) Consistency check.} Only those intermediate patterns $\pi$ are retained for which both $(B \rightarrow C)$ and $B \rightarrow \pi \rightarrow C$ satisfy the required frequency constraints. This ensures semantic consistency. \textbf{5) Transformation.} In the transformation phase, the selected patterns are applied to traces in the original log $\mathcal{L}$. For insertion, if a trace $\sigma \in \mathcal{L}$ contains a frequent direct follower $(B \rightarrow C)$, it may be replaced with the extended sequence $B \rightarrow \pi \rightarrow C$ for some eligible intermediate sequence $\pi$. Conversely, for deletion, if $\sigma$ contains a subsequence of the form $B \rightarrow \pi \rightarrow C$, this may be shortened to the corresponding direct follower $(B \rightarrow C)$. In both cases, the resulting traces preserve the behavioral structure derived from statistically grounded patterns observed in the training log.\\

{\textit{StatisticalReplacement} generalizes process behavior by identifying interchangeable subsequences between shared start and end points, targeting XOR-like structures. \textbf{1) Activity filtering.} The log is filtered as in the \textit{StatisticalInsertion} and \textit{StatisticalDeletion}. \textbf{2) Extract XOR-structures.} Next, we extract frequent patterns of the form $D \rightarrow \rho_i \rightarrow E$, where $\rho_i = \langle A_1, \dots, A_k \rangle$ is an intermediate subsequence of at most $\lambda_{\max}$ activities and each such subsequence must appear in equally or more than factor $\delta$ of traces. \textbf{3) Construct replacement set.} For each fixed pair of start and end points $(D, E)$, the various intermediate alternatives are collected into a replacement set $\mathcal{R}_{(D,E)} = \{ \rho_1, \rho_2, \dots, \rho_n \}$. \textbf{4) Transformation.} During augmentation, if a trace $\sigma \in \mathcal{L}$ contains a segment matching $D \rightarrow \rho_i \rightarrow E$, the intermediate part $\rho_i$ may be substituted with a different alternative $\rho_j \in \mathcal{R}_{(D,E)}$, provided $\rho_j \neq \rho_i$.

For instance, in our running example, \textit{StatisticalInsertion} might add a frequent step $CheckDocs$ between $Order$ and $Pack$, while \textit{StatisticalReplacement} could substitute one observed path $Pack \rightarrow Ship \rightarrow Deliver$ with a special or quick delivery path $Pack \rightarrow ExpressShip \rightarrow Deliver$.

\subsection{Data Augmentation Strategy}\label{subsec:data_augmentation_strategy}
To generate diverse yet semantically coherent process trace variants, we apply two successive augmentations to each training prefix using a pool of transformation functions $\mathcal{Z}$. Our pipeline consists of two core stages: i) \textbf{applicability filtering} and a ii) \textbf{pairwise augmentation check}. Given a prefix of a trace $x$, we first determine which augmentors are applicable by evaluating each method's structural constraints. Our main augmentors (\textit{StatisticalInsertion}, \textit{StatisticalDeletion} and \textit{StatisticalReplacement}) are applied when the sequence matches known behavioral patterns mined from the training log (cf. Section~\ref{subsec:novel_transformation_methods}). If none are applicable, we fall back to simpler, structure-agnostic augmentors named \textit{RandomInsertion}, \textit{RandomDeletion} and \textit{RandomReplacement}. For each input prefix $x$, we record all augmentors that are valid candidates for transformation. Next, we generate two augmented views \( v = t(x), v' = t'(x) \), where \( t, t' \in \mathcal{Z} \) are transformation functions sampled from the augmentation pool \( \mathcal{Z} \), such that \( v \neq v' \).
 We first sample an applicable augmentor uniformly and apply it to produce $v$. Then, we attempt up to 30 trials to generate a second distinct augmentation $v'$, avoiding trivial duplicates. Each augmented pair is padded using left-padding to a fixed length determined by the longest transformed sequence in the batch. This ensures input consistency for self-supervised learning. This strategy ensures each trace contributes semantically meaningful variation to the training set, leveraging both structural knowledge (from frequent patterns) and generalization capability (via fallback augmentors).

 For instance, in our running example, the prefix $Order \rightarrow Pack$ allows only insertion as valid transformation. Among the eligible candidates (derived from \textit{StatisticalInsertion}), both $CheckDocs$ and $ValidatePayment$ can be inserted, yielding $Order \rightarrow CheckDocs \rightarrow Pack$ and $Order \rightarrow ValidatePayment \rightarrow Pack$ as the two augmented prefixes. This illustrates how the strategy selects multiple valid options to generate distinct yet semantically consistent training views.

\subsection{Siamese Pre-Training}\label{subsec:siamese_pre_training}

\begin{figure}[h]
    \centering
    \includegraphics[width=1.0\linewidth]{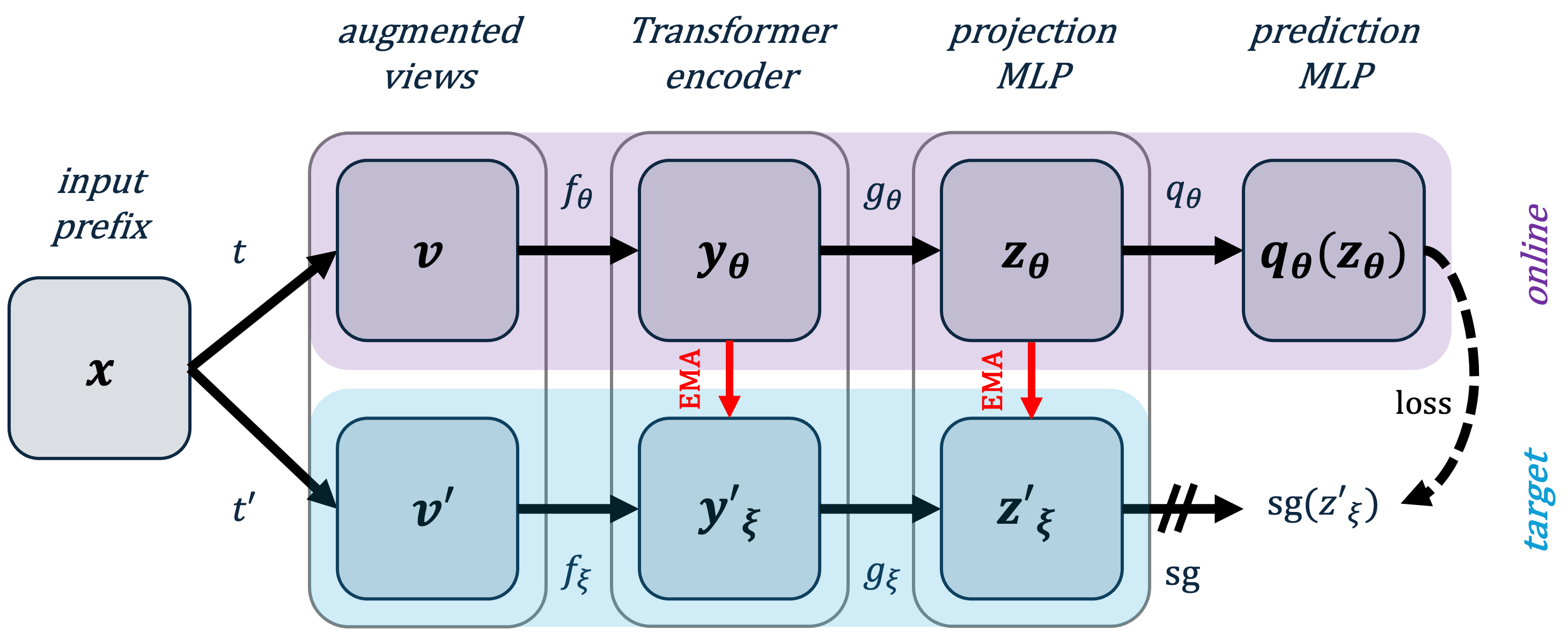}
    \caption{Overview of the pretraining architecture for SiamSA-PPM. Given an input prefix \( x \), two augmented views \( v \) and \( v' \) are generated using transformation functions $t$ and $t'$ chosen following our augmentation strategy and passed through a shared Transformer encoder and projection Multi-Layer Perceptron (MLP). The online network produces a prediction \( q_\theta(z_\theta) \), while the target network (updated via EMA) produces \( z'_\xi \). A loss is computed between the prediction and the stop-gradient of the target to enable representation alignment.}
    \label{fig:byol_architecture}
\end{figure}

To enable effective representation learning for process prefixes without requiring supervision, we adopt a Siamese learning framework inspired by BYOL~\cite{grill2020byol}. This self-supervised strategy facilitates the extraction of high-quality embeddings that capture structural aspects of process behavior. As illustrated in Fig.~\ref{fig:byol_architecture}, the training architecture consists of two neural branches: the \textit{online} network (top) and the \textit{target} network (bottom). Both receive different augmented views $v = t(x)$ and $v' = t'(x)$ of the same input prefix $x$, generated according to our data augmentation strategy (cf. Section \ref{subsec:data_augmentation_strategy}). These views are passed through a shared Transformer encoder backbone:

\begin{equation}
    y_\theta = f_\theta(v), \quad y'_\xi = f_\xi(v')
\end{equation}

Each encoded representation is then mapped to a latent space via a projection MLP: 

\begin{equation}
    z_\theta = g_\theta(y_\theta), \quad z'_\xi = g_\xi(y'_\xi)
\end{equation}

The online branch continues with a prediction MLP $q_\theta$, which is a small feedforward network that maps the projected representation $z_\theta$ to a predicted embedding $\hat{z}_\theta = q_\theta(z_\theta)$. This predictor is essential to avoid representational collapse: it allows the online network to learn a transformation of its own embeddings that can match the slowly evolving target embeddings. The target branch output $z'_\xi$ is detached from the computational graph using the stop-gradient operator $\operatorname{sg}(\cdot)$ and the loss is computed as the similarity between the predicted and target projections:

\begin{equation}
    \mathcal{L}_{\theta,\xi} = 2 - 2 \cdot \frac{q_\theta(z_\theta) \cdot \operatorname{sg}(z'_\xi)}{\|q_\theta(z_\theta)\|_2 \cdot \|\operatorname{sg}(z'_\xi)\|_2}
\label{loss_equation}
\end{equation}

We symmetrize the loss $\mathcal{L}_{\theta,\xi}$ in Eq.~\ref{loss_equation} by separately feeding $v$ to the online network and $v'$ to the target network to compute both $\mathcal{L}_{\theta,\xi}$ and $\tilde{\mathcal{L}}_{\theta,\xi}$. The final training objective is the symmetric loss:

\begin{equation}
\mathcal{L}_{\theta,\xi} = \mathcal{L}_{\theta,\xi} + \tilde{\mathcal{L}}_{\theta,\xi}
\end{equation}

where each term compares the online network's prediction for one view to the target network's projection of the other. At each training step, we perform a stochastic optimization step to minimize $\mathcal{L}_{\theta,\xi}$ with respect to $\theta$ only, but \emph{not} $\xi$.

To avoid collapse (trivial solutions), \cite{chen2021simsiam} claims that only the stop-gradient operation is enough. However, our testing suggests otherwise and collapses without the use of a momentum encoder (introduced in \cite{grill2020byol}). Hence, the target network parameters $\xi$ are not updated by gradient descent but rather using an exponential moving average (EMA) of the online network parameters: 

\begin{equation}
    \xi \leftarrow \tau \cdot \xi + (1 - \tau) \cdot \theta
\end{equation}

where $\tau \in [0, 1)$ is the momentum coefficient. This setup ensures that the models learns to align the online representation with a slowly evolving target representation, allowing robust and stable learning even in the absence of negative samples.

\subsection{Fine-Tuning}\label{subsec:fine_tuning}
Following pre-training, we discard the projection and prediction MLP, but  retain the encoder $f_\theta$ on downstream prediction tasks using labeled data, i.e. next activity prediction and final outcome prediction. We append a softmax classification layer to the encoder output to predict the next event or final outcome. The entire model, including the encoder, is fine-tuned end-to-end using cross-entropy loss. This allows the model to adapt its representations to the specific tasks while benefiting from the robust structure learned during pre-training. After fine-tuning, we evaluate on the trained model.

\section{Experimental Setup}\label{sec:experiments}
In this section we introduce the datasets, metrics, competitors and implementation for our experimental setup.

\subsubsection{Datasets} We evaluate our approach on eight publicly available real-life case studies widely used in the process mining community \footnote{The datasets are available at \protect\url{https://data.4tu.nl}}. \textit{BPIC 2012} contains personal loan and overdraft applications from a Dutch financial institution. It includes three intertwined subprocesses within the same event log, offering rich and realistic financial workflows. \textit{BPIC 2013} consists of IT service management logs from Volvo IT. We use two subsets: \textit{BPIC 2013-c} for closed problems and \textit{BPIC 2013-i} for all incidents. \textit{BPIC 2015} logs building permit applications from five Dutch municipalities. We focus on the first subset, which contains highly variable traces and numerous activity classes. \textit{Sepsis} records emergency department processes from a Dutch hospital involving suspected sepsis cases. \textit{BPIC 2017} pertains to an event log of a Dutch financial institution between 2016 and 2017. Since the model contains no concept drifts, only the first 200,000 events were considered. \textit{Helpdesk} originates from an Italian software company's customer support system. Lastly, we evaluate on the \textit{Bank Account Closure (BAC)} dataset, which is a log referring to a process of an Italian Bank Institution that deals with the closures of bank accounts \footnote{The BAC dataset is available at \protect\url{https://github.com/IBM/processmining/tree/main/Datasets_usecases}}. The dataset contains 32,429 cases with 212,721 events, divided over 15 distinct activities.

\subsubsection{Metrics} Model performance is evaluated using four key metrics. \textit{Running time} measures the time taken for inference by computing the difference between the start and end timestamps during inference. \textit{Average accuracy} assesses overall predictive performance across the dataset: \(\text{Accuracy} = \frac{1}{\mathcal{N}} \sum_{j=1}^{\mathcal{N}} 1\{\hat{y}_j = y_j\}\), where \(\mathcal{N}\) is the total number of labels, \(\hat{y}_j\) the predicted and \(y_j\) the actual label, defined following the prediction problems introduced in Section~\ref{sec:problem_definition}. To assess the variability in augmented event logs, we adopt two entropy-based metrics introduced by \cite{back2019entropy}: \textit{trace entropy} and \textit{prefix entropy}. Trace entropy captures uncertainty at the level of complete traces, measuring how frequently each unique trace occurs. In contrast, prefix entropy considers the diversity of partial traces (prefixes), offering a finer-grained view of process variability by accounting for the workflow order of traces. For next activity prediction, we report the average accuracy across the dataset. For final outcome prediction, we use binary classification targets and evaluate accuracy based on whether the predicted final outcome matches the actual label.

\subsubsection{Baselines} We compare our approach with four state-of-the-art methods for predictive process monitoring. \textit{Tax et al.} \cite{tax2017predictive} use an LSTM-based model where each event is encoded as a feature vector including activity, time since the last event, time of day and weekday. The final hidden state of LSTM summarizes the prefix and is passed to a dense layer to predict the next activity via softmax. \textit{Di Mauro et al.} \cite{dimauro2019activity} propose an inception-based 1D CNN that encodes activity and temporal features into a sequence. Inception modules apply multiple convolutions and pooling in parallel to capture varying local patterns, followed by global max pooling and classification. \textit{Pasquadibisceglie et al.} \cite{pasquadibisceglie2019cnn} transform prefixes into 2D matrices representing control-flow and time-performance data, processed by a standard CNN. This spatial encoding captures temporal patterns without recurrent structures, enabling efficient classification of future activities. \textit{Bukhsh et al.} \cite{bukhsh2021processtransformer} replace recurrence with a transformer encoder using self-attention to model dependencies across the trace. Learnable embeddings with positional encodings are processed through multi-head attention, followed by global pooling and dense layers for prediction.

\subsubsection{Implementation Details} The implementation including the model parameters is available on GitHub \footnote{\protect
\url{https://github.com/SvStraten/SiamSA-PPM}}. Experiments have been performed in Python 3.12.2 using TensorFlow 2.15. We used a temporal train-validation-test split of 65-15-20. To ensure fair comparison, each model is trained and evaluated over five independent repetitions, and we did not run any experiments in parallel. We performed sensitivity analysis on the \textit{BPIC13-c} and \textit{BPIC13-i} datasets. After testing values in the range $10^{-2}$ to $10^{-5}$, we selected $10^{-4}$ as the optimal value for $\alpha$, $\beta$, $\gamma$, and $\delta$. For $\lambda_{\text{max}}$, we have tested integers between 2 and 5, after which 4 has been chosen as the optimal value.


\section{Evaluation Results}\label{sec:results}}
In this section we report results related to predictive accuracy measured under the lens of the metrics introduced in Section~\ref{sec:experiments}.\\ 

\subsubsection{Entropy} 
\begin{figure}[h]
    \centering
    \includegraphics[width=1\linewidth]{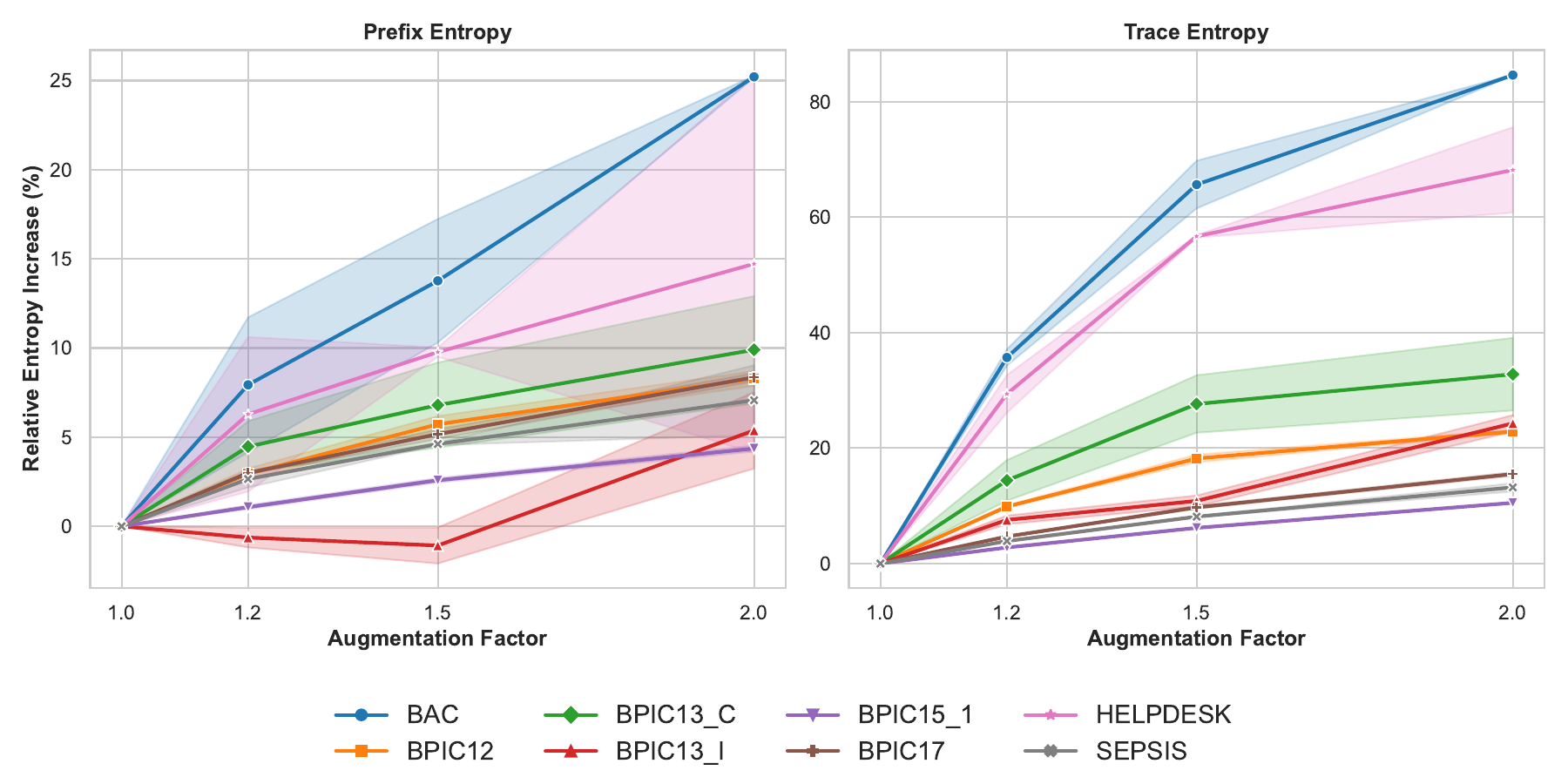}
    \caption{Increase in prefix and trace entropy as a function of augmentation factor, showing how data augmentation enhances variability across different event logs.}
    \label{fig:entropy}
\end{figure}

Fig.~\ref{fig:entropy} illustrates the impact of data augmentation on log variability across different case studies using two entropy-based metrics: prefix entropy (left) and trace entropy (right). These metrics measure the relative increase in variability introduced by our augmentation strategy compared to the original event log on three different augmentation factors: 1.0 (no augmentation), 1.2, 1.5 and 2.0. As the augmentation factor increases from 1.0 to 2.0, most datasets show a consistent rise in both entropy metrics. Notably, the BAC and Helpdesk datasets exhibit the largest gains, with BAC reaching up to a 25\% increase in prefix entropy and over 80\% in trace entropy. These results support our core hypothesis that semantic-preserving augmentation enhances process variability, which is crucial for effective self-supervised PPM.

\subsubsection{Effect of Augmentation on Performance}
We assess the impact of event log augmentation on next activity prediction across five datasets: \textit{Sepsis}, \textit{BPIC15-1}, \textit{BPIC13-i}, \textit{BPIC13-c}, and \textit{Helpdesk}. Each baseline model is evaluated on the original (base) logs, logs augmented using the method in \cite{kappel2023augmentation} and logs augmented with our proposed strategy (cf. Section~\ref{subsec:data_augmentation_strategy}). Table~\ref{tab:augmented_logs} reports results using the best-performing augmentation factor (1.2, 1.5 or 2.0). On \textit{BPIC15-1}, our method increases performance for all baselines, while the augmentation strategy in \cite{kappel2023augmentation} does not consistently yield improvements. For the other datasets, both methods typically lead to gains over the base logs, with most models showing higher accuracy when trained on augmented data. Notably, on \textit{Sepsis} and \textit{BPIC13-i} almost all baselines benefit from augmentation. These results confirm that augmentation, whether via our approach or prior work, can enhance predictive performance, though the effectiveness varies by dataset and model.

\begin{table}[h]
\centering
\caption{Accuracy of the baselines for next activity prediction using base event log, the augmented log following \cite{kappel2023augmentation} and the augmented log following our method (cf. Section \ref{subsec:data_augmentation_strategy}).}
\label{tab:augmented_logs}
\renewcommand{\arraystretch}{1.2}
\resizebox{\columnwidth}{!}{%
\small
\begin{tabular}{|l|l|c|c|c|c|}
\hline
\textbf{Dataset} & \textbf{Log Type} 
  & \textbf{Tax \cite{tax2017predictive}} 
  & \textbf{Di\,Mauro \cite{dimauro2019activity}} 
  & \textbf{Pasquad.\ \cite{pasquadibisceglie2019cnn}} 
  & \textbf{Bukhsh \cite{bukhsh2021processtransformer}}\\
\hline

\multirow{3}{*}{Sepsis} 
  & Baseline                & $52.69 \pm 1.40$ & $59.44 \pm 1.93$ & $56.07 \pm 1.72$ & $59.16 \pm 2.18$ \\
  & + \cite{kappel2023augmentation} & $52.89 \pm 1.81$ & $59.83 \pm 2.11$ & $55.93 \pm 2.97$ & $60.29 \pm 1.65$ \\
  & + Our Augment.          & $\bm{52.91} \pm \bm{1.09}$ & $\bm{59.87} \pm \bm{1.27}$ & $\bm{56.39} \pm \bm{1.54}$ & $\bm{61.50} \pm \bm{1.12}$ \\
\hline

\multirow{3}{*}{BPIC15-1} 
  & Baseline                & $24.83 \pm 0.44$ & $47.15 \pm 0.55$ & $41.90 \pm 0.46$ & $38.54 \pm 1.53$ \\
  & + \cite{kappel2023augmentation} & $22.12 \pm 3.44$ & $44.12 \pm 1.58$ & $42.31 \pm 1.21$ & $37.48 \pm 1.65$ \\
  & + Our Augment.          & $\bm{29.34} \pm \bm{1.62}$ & $\bm{47.46} \pm \bm{0.36}$ & $\bm{42.78} \pm \bm{0.89}$ & $\bm{40.49} \pm \bm{0.55}$ \\
\hline

\multirow{3}{*}{BPIC13-i} 
  & Baseline                & $55.02 \pm 0.51$ & $59.39 \pm 0.11$ & $\bm{59.10} \pm \bm{3.79}$ & $58.69 \pm 0.12$ \\
  & + \cite{kappel2023augmentation} & $\bm{56.45} \pm \bm{1.01}$ & $59.98 \pm 1.67$ & $58.56 \pm 3.67$ & $57.96 \pm 1.32$ \\
  & + Our Augment.          & $55.34 \pm 0.22$ & $\bm{61.24} \pm \bm{0.88}$ & $59.02 \pm 2.78$ & $\bm{58.98} \pm \bm{0.28}$ \\
\hline

\multirow{3}{*}{BPIC13-c} 
  & Baseline                & $\bm{55.18} \pm \bm{3.17}$ & $56.85 \pm 2.01$ & $\bm{52.31} \pm \bm{0.17}$ & $55.92 \pm 2.23$ \\
  & + \cite{kappel2023augmentation} & $53.81 \pm 3.73$ & $57.89 \pm 1.97$ & $52.01 \pm 2.56$ & $54.97 \pm 2.69$ \\
  & + Our Augment.          & $54.20 \pm 1.26$ & $\bm{58.19} \pm \bm{2.10}$ & $45.69 \pm 4.52$ & $\bm{56.72} \pm \bm{1.26}$ \\
\hline

\multirow{3}{*}{Helpdesk} 
  & Baseline & $\bm{70.63} \pm \bm{0.44}$ & $\bm{78.85} \pm \bm{0.04}$ & $\bm{78.81} \pm \bm{0.17}$ & $78.45 \pm 0.25$ \\
  & + \cite{kappel2023augmentation} & $69.21 \pm 0.68$ & $75.56 \pm 0.43$ & $78.64 \pm 0.32$ & $78.76 \pm 0.17$ \\
  & + Our Augment. & $68.79 \pm 0.78$ & $77.54 \pm 0.56$ & $78.71 \pm 0.12$ & $\bm{78.93} \pm \bm{0.10}$ \\
\hline
\end{tabular}%
}
\end{table}

\subsubsection{Next Activity Prediction} Table~\ref{tab:next_activity} reports the mean prediction accuracy for next activity prediction across our selected datasets. SiamSA-PPM achieves competitive performance compared to the state-of-the-art baselines, demonstrating especially strong results on low-variance and structured datasets. In particular, our model obtains the best results on BPIC13-c (57.06\%), Sepsis (60.26\%) and BAC (95.49\%), outperforming all other approaches. Although our accuracy on some large datasets (e.g., BPIC12 and BPIC17) is slightly lower than the best-performing baselines, our approach remains competitive. In addition, on the highly imbalanced BPIC15-1 dataset, our model still remains competitive despite the dataset's challenging nature with 298 activity classes.

\begin{table}[h]
\centering
\caption{Accuracy (\textbf{mean} $\pm$ \textbf{std}) for Next Activity Prediction. \textbf{Bold} denotes the highest score, \textit{italic} and \underline{underline} the second and third highest respectively.}
\label{tab:next_activity}
\renewcommand{\arraystretch}{1.3}
\resizebox{\columnwidth}{!}{%
\small
\begin{tabular}{|l|c|c|c|c|c|}
\hline
\textbf{Dataset} 
  & \textbf{Tax \cite{tax2017predictive}} 
  & \textbf{Di Mauro \cite{dimauro2019activity}} 
  & \textbf{Pasquad. \cite{pasquadibisceglie2019cnn}} 
  & \textbf{Bukhsh \cite{bukhsh2021processtransformer}} 
  & \textbf{SiamSA-PPM} \\
\hline
BPIC12         
  & 70.85 $\pm$ 0.79     
  & \textit{83.20} $\pm$ \textit{1.00}              
  & \underline{83.14} $\pm$ \underline{0.27}      
  & $\bm{83.47} \pm \bm{0.31}$      
  & 79.48 $\pm$ 0.05 \\
  \hline
BPIC13-c       
  & 55.18 $\pm$ 3.17     
  & \textit{56.85} $\pm$ \textit{2.01}             
  & 52.31 $\pm$ 0.17      
  & \underline{55.92} $\pm$ \underline{2.23}      
  & $\bm{57.06} \pm \bm{0.24}$ \\
  \hline
BPIC13-i       
  & 55.02 $\pm$ 0.51     
  & $\bm{59.39} \pm \bm{0.11}$              
  & \underline{59.10} $\pm$ \underline{3.79}      
  & 58.69 $\pm$ 0.12      
  & \textit{59.27} $\pm$ \textit{0.08} \\
  \hline
BPIC15-1       
  & 24.83 $\pm$ 0.44     
  & $\bm{47.15} \pm \bm{0.55}$              
  & \textit{41.90} $\pm$ \textit{0.46}      
  & \underline{38.54} $\pm$ \underline{1.53}      
  & 37.11 $\pm$ 0.13 \\
  \hline
BPIC17         
  & 75.25 $\pm$ 0.11 
  & $\bm{88.77} \pm \bm{0.13}$ 
  & \underline{88.48} $\pm$ \underline{0.23} 
  & \textit{88.63} $\pm$ \textit{0.12} 
  & 85.65 $\pm$ 0.06 \\
  \hline
Sepsis         
  & 52.69 $\pm$ 1.40     
  & \textit{59.44} $\pm$ \textit{1.93}              
  & 56.07 $\pm$ 1.72      
  & \underline{59.16} $\pm$ \underline{2.18}      
  & $\bm{60.26} \pm \bm{0.20}$ \\
  \hline
Helpdesk       
  & 70.63 $\pm$ 0.44     
  & $\bm{78.85} \pm \bm{0.04}$              
  & \textit{78.81} $\pm$ \textit{0.17}      
  & 78.45 $\pm$ 0.25      
  & \underline{78.60} $\pm$ \underline{0.18} \\
  \hline
BAC 
  & 80.32 $\pm$ 0.31    
  & 95.12 $\pm$ 0.20              
  & \underline{95.19} $\pm$ \underline{0.22}       
  & \textit{95.47} $\pm$ \textit{0.03}    
  & $\bm{95.49} \pm \bm{0.01}$  \\
\hline
\end{tabular}%
}
\end{table}


\subsubsection{Final Outcome Prediction} Table~\ref{tab:final_outcome} summarizes the accuracy results for binary classification targets on the BPIC12 and Sepsis datasets on Final Outcome Prediction problem (cf.\ Section~\ref{sec:problem_definition}). It is important to note that not all datasets are equally relevant for this task, as some datasets do not include relevant outcome targets. BPIC13-c, BPIC13-i and Helpdesk  do not contain varying outcomes for instance. For BPIC12, the outcomes correspond to the status of financial application requests: \textit{Approved}, \textit{Declined} and \textit{Cancelled}. Our model consistently ranks near the top on all three targets, reaching 80.04\% accuracy on the \textit{Declined} class, just 0.82\% below the top performer and closely matches the best scores for \textit{Approved} and \textit{Cancelled}. In the Sepsis dataset, the prediction targets indicate the reason for patient release after treatment. These include \textit{Release-A}, \textit{Release-B}, \textit{Release-C} and \textit{Release-D}, which reflect different clinical justifications for discharge. SiamSA-PPM achieves the highest accuracy for \textit{Release-B} (94.21\%) and \textit{Release-D} (96.02\%), and remains competitive on the remaining targets, demonstrating robust performance in distinct clinical outcomes. \\


\begin{table}[ht]
\centering
\caption{Accuracy (\textbf{mean} $\pm$ \textbf{std}) for Final Outcome Prediction.}
\label{tab:final_outcome}
\renewcommand{\arraystretch}{1.3}
\resizebox{\columnwidth}{!}{%
\small
\begin{tabular}{|l|c|c|c|c|c|}
\hline
\textbf{Target} 
  & \textbf{Tax \cite{tax2017predictive}} 
  & \textbf{Di Mauro \cite{dimauro2019activity}} 
  & \textbf{Pasquad. \cite{pasquadibisceglie2019cnn}} 
  & \textbf{Bukhsh \cite{bukhsh2021processtransformer}} 
  & \textbf{SiamSA-PPM} \\
\hline
\hline
\multicolumn{6}{|l|}{\textbf{BPIC12}} \\
Approved     
  & $65.63 \pm 3.40$           
  & $56.42 \pm 0.25$           
  & $\textit{74.73} \pm \textit{0.58}$     
  & $\bm{75.43} \pm \bm{0.22}$     
  & $\underline{73.58} \pm \underline{0.14}$ \\
Declined     
  & $75.31 \pm 1.50$           
  & $69.70 \pm 0.64$           
  & $\underline{79.50} \pm \underline{1.15}$     
  & $\bm{80.86} \pm \bm{0.04}$     
  & $\textit{80.04} \pm \textit{0.02}$ \\
Cancelled    
  & $63.36 \pm 2.48$           
  & $\bm{75.76} \pm \bm{0.01}$           
  & $74.47 \pm 3.02$     
  & $\textit{75.63} \pm \textit{1.13}$     
  & $\underline{75.39} \pm \underline{0.06}$ \\
\hline
\hline
\multicolumn{6}{|l|}{\textbf{Sepsis}} \\
Release-A    
  & $\bm{84.15} \pm \bm{2.27}$           
  & $77.39 \pm 0.98$           
  & $\underline{81.95} \pm \underline{0.48}$     
  & $\textit{83.42} \pm \textit{0.83}$     
  & $80.71 \pm 0.11$ \\
Release-B    
  & $92.47 \pm 1.75$           
  & $91.17 \pm 2.42$           
  & $\underline{93.28} \pm \underline{0.43}$     
  & $\textit{94.06} \pm \textit{0.65}$     
  & $\bm{94.21} \pm \bm{0.08}$ \\
Release-C    
  & $90.58 \pm 0.11$           
  & $90.10 \pm 0.49$           
  & $\textit{90.80} \pm \textit{0.06}$     
  & $\bm{90.82} \pm \bm{0.00}$     
  & $\underline{90.59} \pm \underline{0.00}$ \\
Release-D    
  & $94.94 \pm 1.27$           
  & $\textit{95.81} \pm \textit{0.58}$           
  & $\underline{95.64} \pm \underline{0.78}$     
  & $95.58 \pm 0.72$     
  & $\bm{96.02} \pm \bm{0.11}$ \\
\hline
\end{tabular}%
}
\end{table}


\begin{figure}[h]
    \centering
    \includegraphics[width=1\linewidth]{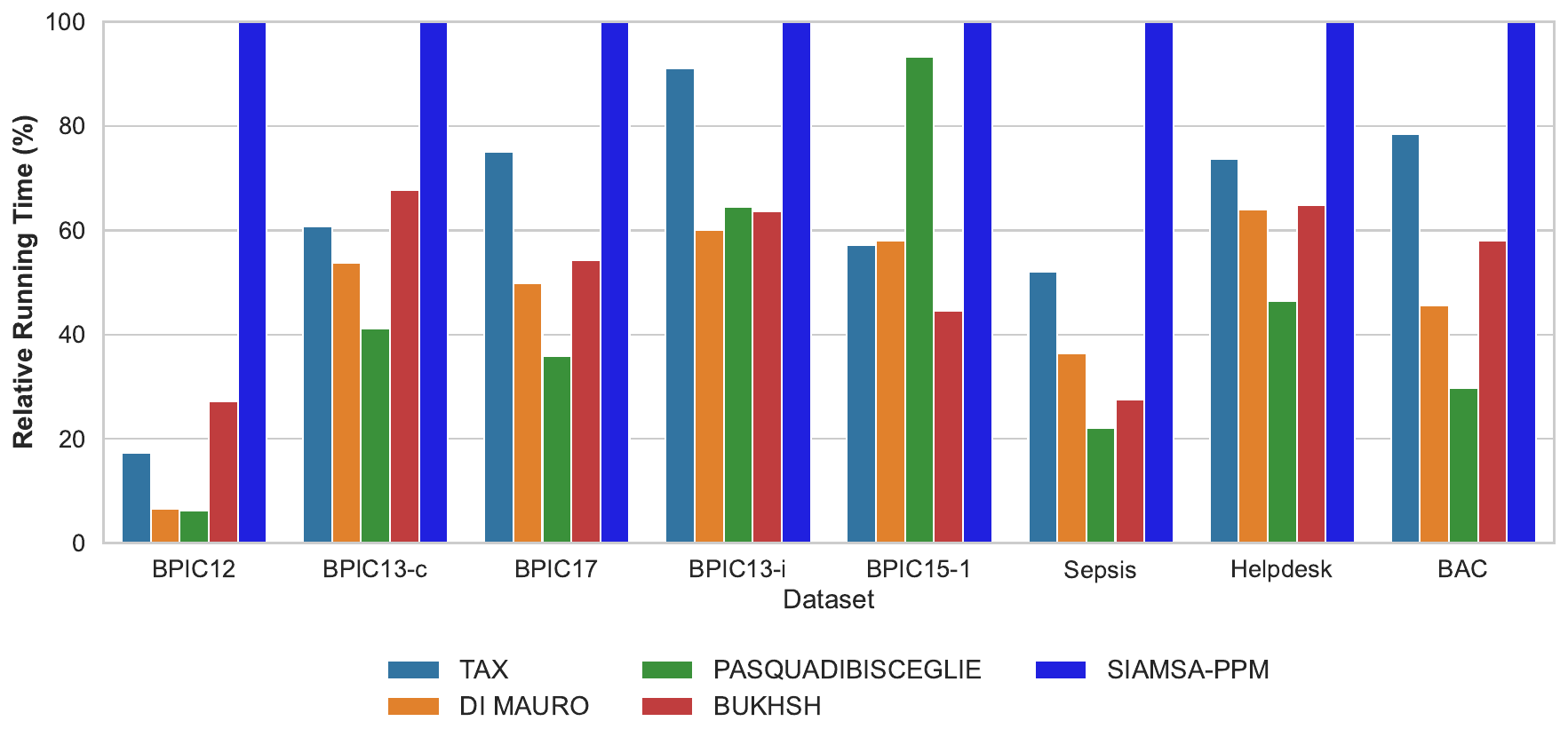}
    \caption{Relative inference time (normalized to the slowest method per dataset) across all benchmark datasets.}
    \label{fig:running_times}
\end{figure}

\subsubsection{Running Time} We evaluate the computational cost of each method by measuring the average inference time per dataset, as shown in Fig.~\ref{fig:running_times}. Our method exhibits a longer inference time compared to competitors, particularly in large or complex datasets such as BPIC12, BPIC17 and Sepsis. 
Our architecture contains approximately 666,000 trainable parameters on the Helpdesk dataset, while the second slowest approach (\textit{Tax et al.}) contains just 123,000 parameters. Since this is more than 5 times the size, the significantly larger capacity leads to increased computational overhead. However, our method trades off computational efficiency for higher model capacity and predictive performance. While inference is slower compared to the baselines, the cost remains acceptable in most practical settings. For applications where predictive accuracy and robustness across diverse datasets are prioritized, our approach offers a strong, albeit more resource-intensive, alternative.\\


\subsubsection{Ablation Study} To evaluate the effectiveness of our data augmentation strategy, we conducted an ablation study on the BPIC13-closed dataset. As shown in Table~\ref{tab:accuracy_comparison}, applying pre-training using our random methods (i.e. \textit{RandomInsertion}, \textit{RandomDeletion} and \textit{RandomReplacement}) provides a minor improvement over the supervised-only baseline (56.09\% vs. 55.67\%). In contrast, using our proposed augmentation strategy, described in Section~\ref{subsec:data_augmentation_strategy}, leads to a more notable increase, achieving 57.06\% accuracy. This highlights the importance of targeted augmentation design in optimizing pre-training benefits.

\begin{table}[h]
\centering
\caption{Accuracy comparison for different training strategies.}
\label{tab:accuracy_comparison}
\renewcommand{\arraystretch}{1.2}
\small
\begin{tabular}{|l|c|}
\hline
\textbf{Training Strategy} & \textbf{Test Accuracy (\%)} \\
\hline
Supervised only                & $55.67 \pm 0.21$ \\
\hline
+ Pre-training (random strategy)  & $56.09 \pm 0.28$ \\
\hline
+ Pre-training (our strategy)     & $\bm{57.06} \pm \bm{0.24}$ \\
\hline
\end{tabular}%

\end{table}

\begin{figure}[h]
    \centering
    \includegraphics[width=0.9\linewidth]{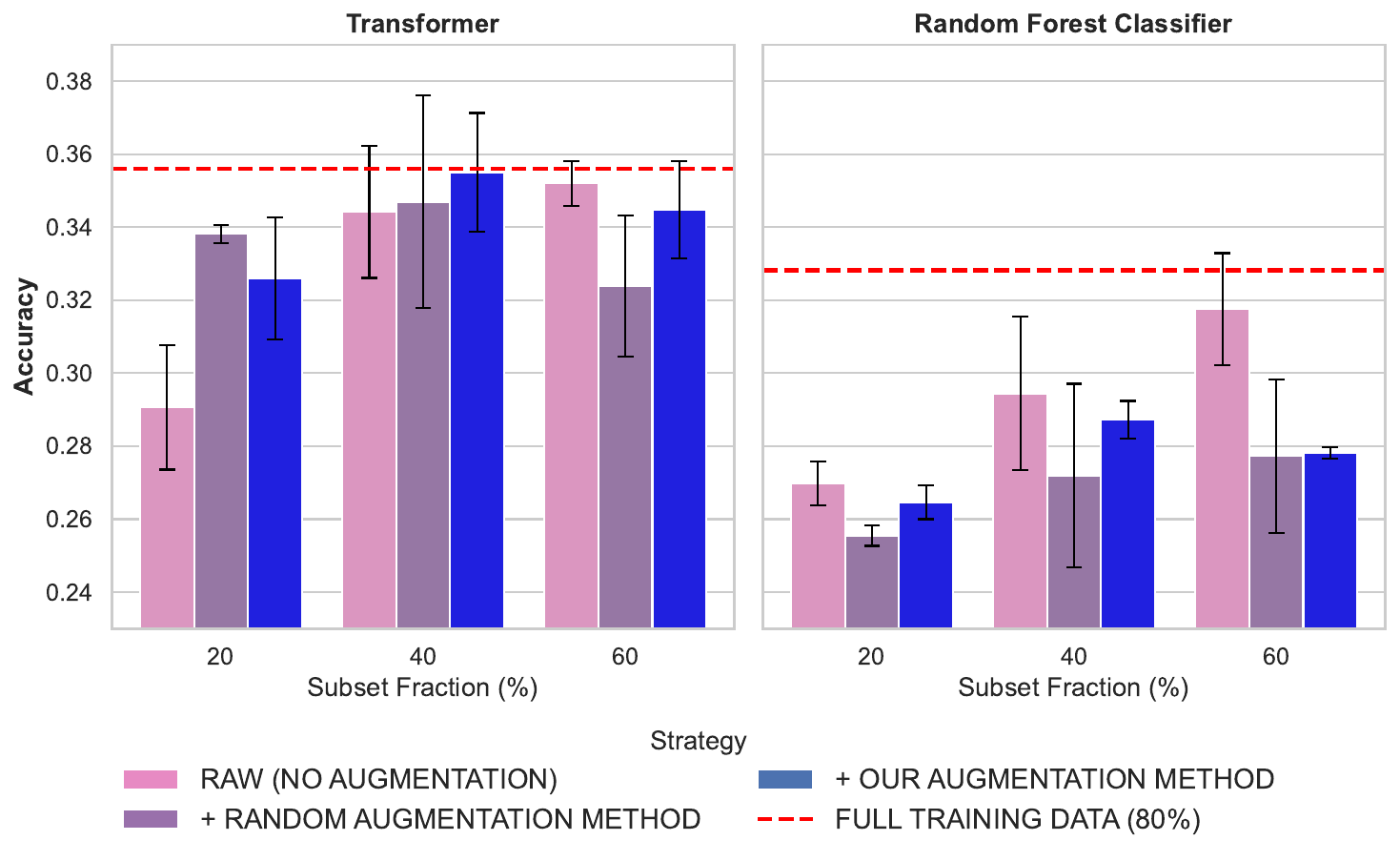}
    \caption{Accuracy comparison between a deep learning Transformer model (left) and shallow Random Forest Classifier (right) on the BPIC 2015-1 dataset, under varying raw training subset sizes (20\%, 40\% and 60\%). Each strategy (no augmentation, random augmentation and our augmentation method) uses the same base data, but the latter two are upsampled to the original 80\% training size. The red dashed line denotes the baseline of training on the full 80\% raw training data.}
    \label{fig:ablation_study}
\end{figure}

Fig. ~\ref{fig:ablation_study} illustrates the impact of data augmentation on model accuracy using varying fractions (20\%, 40\%, 60\%) of the BPIC15-1 training data. We compare a Transformer model and a Random Forest classifier, each trained on raw subsets, subsets augmented using a random strategy and subsets augmented using our proposed method. For the Transformer, our method consistently improves accuracy at 20\% and 40\% sample sizes, bringing performance close to that achieved with the full 80\% training data. Interestingly, on the smallest subset (20\%), random augmentation performs slightly better, likely due to the limited information in the fewer number of traces. In contrast, the Random Forest classifier shows no performance gains with either augmentation method, remaining well below the full-data baseline. These results suggest that deep learning models are better equipped to exploit synthetic variation introduced through augmentation, leading to better performance.

\section{Conclusion and Future Work}\label{sec:conclusion_and_future_work}

In this paper, we introduced a novel framework that combines statistically grounded data augmentation techniques with a self-supervised Siamese learning approach to enhance PPM. By addressing the challenges of limited and low-diversity event log data, we demonstrate that our novel transformation methods, \textit{StatisticalInsertion}, \textit{StatisticalDeletion} and \textit{StatisticalReplacement}, preserve process semantics while enriching trace variability. Coupled with a BYOL-inspired Siamese architecture, SiamSA-PPM learns robust and generalizable representations of process prefixes without reliance on manual labels. Extensive experiments across eight real-life case studies confirmed that SiamSA-PPM achieves competitive or superior performance in both next activity and final outcome prediction tasks. Ablation studies further validated the effectiveness of our tailored augmentation strategies over generic random transformations, highlighting their contribution to model accuracy and data efficiency.

Although our results are promising, several avenues remain for future work. Exploring our approach in a streaming or online continual learning setup is an interesting future direction. Additionally, extending our framework to other PPM tasks such as remaining time prediction or anomaly detection could unlock broader applicability, along with the generation of not only activity sequences, but also associating them with timestamps and other attributes. Last but not least, a future direction of this work involves integrating the proposed data augmentation framework into Prescriptive Process Analytics pipelines. Many such frameworks depend on PPM techniques to identify which ongoing process instances warrant intervention, often relying on outcome prediction models. A promising extension of this work would be to tailor the augmentation strategy specifically to reduce false positives generated by these models. This could, in turn, lower the number of unnecessary interventions, which typically entail significant operational costs for organizations in terms of both time and resources.

%
%
%
%
\bibliographystyle{splncs04}
\bibliography{references}

\end{document}